\documentclass[journal,10pt]{IEEEtran}

\usepackage{booktabs}
\usepackage{graphicx}
\usepackage{subfigure}
\usepackage{slashbox}
\usepackage[ruled]{algorithm2e}
\usepackage{cite}
\makeatletter

\newcommand{\Rmnum}[1]{\expandafter\@slowromancap\romannumeral #1@}
\makeatother
\ifCLASSINFOpdf
\else
\fi

\usepackage{amssymb}

\usepackage{mathrsfs}
\usepackage{amsmath}

\usepackage{graphicx}
\usepackage{caption}
\usepackage{algorithmicx}
\usepackage{algpseudocode}
\usepackage{multirow}
\usepackage{amssymb}
\usepackage{makecell}

\begin{document}
%
\title{Orientation Convolutional Networks \\
	for Image Recognition}

\author{Yalan~Qin,
	Guorui~Feng, Hanzhou~Wu,
	Yanli~Ren,
	and Xinpeng~Zhang
	\thanks{Y. Qin, G. Feng, Y. Ren and X. Zhang are with the School of Communication and Information Engineering, Shanghai University, Shanghai 200444, China. \emph{(Corresponding author: Guorui Feng.)}}}

\markboth{}%
{Shell \MakeLowercase{\textit{et al.}}: Bare Demo of IEEEtran.cls for Journals}

\maketitle

\begin{abstract}
Deep Convolutional Neural Networks (DCNNs) are capable of obtaining powerful image representations, which have attracted great attentions in image recognition. However, they are limited in modeling orientation transformation by the internal mechanism. In this paper, we develop Orientation Convolution Networks (OCNs) for image recognition based on the proposed Landmark Gabor Filters (LGFs) that the robustness of the learned representation against changed of orientation can be enhanced. By modulating the convolutional filter with LGFs, OCNs can be compatible with any existing deep learning networks. LGFs act as a Gabor filter bank achieved by selecting $ p $ $ \left( \ll n\right) $ representative Gabor filters as andmarks and express the original Gabor filters as sparse linear combinations of these landmarks. Specifically, based on a matrix factorization framework, a flexible integration for the local and the global structure of original Gabor filters by sparsity and low-rank constraints is utilized. With the propogation of the low-rank structure, the corresponding sparsity for representation of original Gabor filter bank can be significantly promoted. Experimental results over several benchmarks demonstrate that our method is less sensitive to the orientation and produce higher performance both in accuracy and cost, compared with the existing state-of-art methods. Besides, our OCNs have few parameters to learn and can significantly reduce the complexity of training network.
\end{abstract}


\IEEEpeerreviewmaketitle

\section{INTRODUCTION}
Oriented filters play a key role to extract robust image representation. Hand-crafted features with orientational information encoded like Gabor features \cite{Perona91} have been investigated. The sample rate of orientation interval for Gabor function is expected to be dense enough at each fixed central frequency \cite{GM95}. Such as in \cite{T20}, the sample rate of 16 is adopted in the application of script identification. Recently, Deep Convolutional Neural Networks (DCNNs) have alleviated the effect of local transition by properties of the convolution and max-pooling. Meanwhile, the inherent capability of dealing with large object transformation is limited, resulting in poor performance in image classification, oriented object detection, and many other visual tasks \cite{YTang19,HXiao12}.

Due to the lack of capacity to transformation, the direct way is to design the filter in a proper way. In early years, a general architecture was proposed to synthesize filters of arbitrary orientations based on linear combinations of base filters. It enables one to steer a filter to any orientations adaptively, and determine the output of the filter as a function of the orientation \cite{Freeman91}. However, these filters are typically based on directional derivatives and their symmetry produces orientational responses with period $ \pi $ and independent of image structure \cite{Simoncelli96}. Currently, steerable CNNs \cite{Taco16} present a theoretical framework for understanding the steerable representation in the convolutional network, and show that steerability is a useful inductive bias for improving the model accuracy. Actively rotating filters (ARFs) enforce DCNNs to have the generalization capacity of the rotation \cite{Zhou17}. However, such rotation for the filter is not proper for large and complex filters. Gabor convolutional networks (GCNs) \cite{S18} are proposed to give deep features the capacity of dealing with orientation and scale based on Gabor filters. However, such manipulation is limited to the discriminating capability for Gabor filters with fixed orientations and scales.

In last decades, matrix factorization  \cite{Lee99} has become very hot for the data representation. It compresses data by seeking for the set of basis vectors and the corresponding representation for the data Sparse coding \cite{2003Optimally,96Ol} method is proposed as the matrix factorization and has been shown useful for real applications in different areas such as image restoration \cite{06Elad,08JMairal}, salient object detection \cite{YZ19,GM20} and visual neuroscience \cite{08Lee}. Since the data can be encoded by sparse representations with just a few coefficients, the computational cost can be reduced and the achieved coding result is relatively easy to explain. Different sparse based algorithms have been proposed, \emph{i.e.,} sparse nonnegative matrix factorization \cite{Hoyer04}, sparse principle component analysis \cite{2019Sparse_Prin}. Moreover, a variety of researches have been focused on optimization algorithms for the sparse coding \cite{10Gregor,07BSch}.

Several studies have considered exploring both the local and global structures simultaneously \cite{Y.X13,V.M15}. The structured representation such as the sparse and low-rank representation is evaluated over transformed data points. Although these methods have achieved the state-of-art performances in various subspace clustering applications \cite{2021Sparse}, a major problem of these existing methods is the representation ability for one latent learned subspace to data points which may lie in a union of subspaces. Furthermore, no explicit theoretical proofs support that the final structured representation corresponds to the correct clustering structure. To this end, we need to seek for an efficient integration strategy for the local and global structure.

As is known, the convolution is a linear operation. Since the steerable filter can synthesize filters of arbitrary orientations by the combinations of different basis filters in a linear way. If a filter is steerable regarding the rotation, we can obtain its output corresponding to the rotated version by linearly combining outputs of filters. If the number of basis filters is smaller than the total number of orientation samples to be filtered, both the computation and the storage will be saved. For simplicity, we can assume that original oriented filters are sampled from a union of different subspaces which are independent to achieve the above effect. 

Based on the above observation, we propose Landmark Gabor Filters (LGFs) and then apply LGFs to extend the DCNNs to develop a new deep learning architecture termed Orientation Convolutional Networks (OCNs) for image recognition. Specifically, LGFs generates $ p $ $ \left( \ll n\right) $ representative Gabor filters as landmarks and the remaining Gabor filters are represented as sparse combinations of the landmarks in a linear way. The basic idea of OCNs is simple but effective. It can effectively capture the orientation changes by using LGFs to manipulate the convolutional filter in DCNNs. Unlike most existing DCNNs methods, our method simultaneously considers learning transformations which are parameterized with a neural network and achieving an effectively oriented filter bank. Compared with traditional approaches, OCNs is a deep model, which can map samples with orientation changes into the latent space explicitly, with parameters learned by a data-driven manner. 

The contribution of our work is twofold. From the perspective of pure oriented filter bank, we focus on how to let it benefit from the merits of DCNNs where the nonlinear feature representation can be obtained. From the perspective of deep neural networks, we analyse that integrating the advantages of effectively oriented filter bank and deep learning for reinforcing the robustness of the representation against orientation changes is feasible .
\section{Related Work}
\subsection{Steerable Filters and Gabor Filters}
Steerable filter is adoted to express a kind of filters where a filter for arbitrary orientation can be obtained using a linear combination of the set of basis filters \cite{Freeman91}. And the function of rotation transform$ f(x,y):R^{2} \rightarrow C $ is steerable regarding the rotation:\\
\begin{equation}
f^{\theta}(x,y)=\sum\limits_{j=1}^{M}k_{j}(\theta)\varphi^{j}(x,y), \label{XX}
\end{equation}
where $ f^{\theta}(x,y) $ denotes  the rotated version of $ f(x,y) $ by the angle $ \theta $. $ \varphi^{j}(x,y) $ is the the base function independent of the rotation angle $ \theta $, and $ k_{j}(\theta) $ is the steering function of $ f(x,y) $. Streerable CNNs \cite{Taco16} present a theoretical framework for understanding the steerable representation in convolutional network, and show that steerability is a useful inductive bias for improving model accuracy.

Gabor wavelets are proposed by utilizing complex functions to act as the basis for related Fourier transforms \cite{46Gabor}. Gabor filters are widely adopted for modeling receptive fields of simple cells of the visual cortex, and are as follows \cite{10Baochang}:
\begin{equation}
G(u,v)=\frac{||k_{u,v}||^{2}}{\sigma^{2}}e^{-(||k_{u,v}||^{2}||z||^{2}/2\sigma^{2})}\left[ e^{ik_{u,v}z}-e^{-\sigma^{2}/2}\right],
\end{equation}
where $ u $ denotes the orientation, $ v $ is the frequency, $ k_{u}=u\frac{\pi}{U} $, $ k_{v}=(\pi/2)/\sqrt{2}(v-1) $, with $ u=0,1,...,U $ and $ v=0,1,...,V $, $ \sigma=2\pi $ and $ z=(x,y) $.
\subsection{Matrix Factorization and Sparse Coding}
Matrix factorization is quite hot for data representation. For each data point, it aims for compressing the data by seeking for a set of basis vectors and the representation corresponding to the related basis. Given the data matrix $ X=\left[ x_{1}, x_{2},...,x_{n}\right]\in R^{m\times n}$, matrix factorization is defined by seeking for matrices $ U\in R^{m\times p} $ and $ V\in R^{p\times n} $ whose product can well approximate $ X $ \cite{HL14}:
\begin{equation}
X\approx UV,
\end{equation}
where each column of $ U $ can be seen the basis vector and each column of $ V $ is the representation corresponding to the basis vector. Frobenius norm of a matrix  $ ||\;.\;||_{F}^{2} $ can be used to measure the approximation. Then, it can be formulated as:
 \begin{equation}
 \min_{U,V}||X-UV^{T}||_{F}^{2}.
 \end{equation}
Obviously, the solution $ V $ of the optimization problem is inevitablely non-sparse. To address this issue, sparse coding  \cite{Hoyer04} is proposed as a matrix factorization technique for encoding different data points with only a few coefficients. Sparse coding adds the sparse constraint on $ V $ in the optimization (3) and is defined as follows
\begin{equation}
\min_{U,V}||X-UV||_{F}^{2}+\alpha ||V||_{0}
\label{XX}
\end{equation}
where $ ||\;.\;||_{0} $ is used for measuring the sparsity  and $ \alpha $ denotes a coefficient controlling the sparsity penalty. By imposing the low-rank structure on $ V $ in (3), the low-rank representation is defined as follows
\begin{equation}
\min_{U,V}||X-UV||_{F}^{2}+\beta rank(V)
\label{XX}
\end{equation}
where $ rank(V) $ denotes the rank value of $ V $ and $ \beta $ is a coefficient controlling the low-rank penalty. Under the low-rank constraint, the correlation between different data points is enhanced within clusters and weakened for different clusters. However, the low-rank representation is not guaranteed to be of the block-diagonal structure \cite{J.F14} in practice, which reflects the expected relationship of data points. Besides, the low-rank representation is not able to fully utilize the local information.
\subsection{Robust features learning}
Given rich, and redundant, convolutional filters, data augmentation can be used to obtain local or global transform invariance \cite{DA01}. Despite the performance of the data augmentation, the main drawback is that learning all possible transformations often needs more parameters, resulting in the increasing training cost and the risk of over-fitting significantly. Recently, TI-Pooling \cite{Laptev16} is proposed to use the parallel network architectures for the considered transform set. Nevertheless, with the data augmentation, TI-Pooling requires dramatically more cost for training and testing than the standard DCNNs.

\textbf{Deformable convolutional networks:} Deformable convolution is proposed in \cite{Wei17} to greatly increase the transformation modeling capability of CNNs, enabling the network robust to dense spatial transformations in a simple, deep, efficient and end-to-end manner. However, small size filters are preferred for the deformable filters, \emph{i.e.,} $ 3\times 3 $.

\textbf{Spatial Transform Network:} Spatial Transformer Network (STN) \cite{Jaderberg15} uses an additional network module to manipulate the feature maps. STN indeed proposes a general framework for the spatial transform, but fails to estimate the complex transform parameters precisely by giving a solution. 

\textbf{Wide Residual Networks:} Wide Residual Networks (WRN) \cite{Zagoruyko16} performs a detailed experimental investigation on the ResNet \cite{He06} networks as well as proposes a novel architecture to decrease the depth and increase the width of the ResNet networks. However, the size of parameters for WRN is relatively large.

\textbf{Oriented Response Networks:} Oriented Response Network (ORN) \cite{Zhou17} is introduced to actively rotate filters (ARFs) during the convolution and produce feature maps with location and orientation encoded. The ARF can be viewed as a virtual filter bank including the filter itself and multiple un-materialised rotated versions corresponding to it. However, ORN is just applicable for filters with small size.

\textbf{Gabor Convolutional Networks:} Gabor Convolutional Networks (GCNs) \cite{S18} incorporates Gabor filters into deep convolutional neural networks (DCNNs). By only manipulating the convolution operator, GCNs can be implemented and are compatible with the current deep learning architecture. Such a modulation is limited to the discriminating capability for orientation and scale using Gabor filters, which is absent of considering some cases Gabor filters cannot capture.

To deal with different rotation angels, we require a considerable amount of oriented filters, \emph{i.e.,} Gabor filters. In practice, due to the orientation interpolation caused by the operation of multi-layer pooling, we can utilize a limited amount of orientations to ensure the accuracy. Furthermore, motivated by the fact that the dimension of basis vector (column vectors of $ U $)  in (3)  is the same as original Gabor filters, we can regard basis vectors as LGFs of original Gabor filters to reduce the complexity as well as guarantee the performance.
 
Different from existing work, we propose a new model to bridge LGFs and deep neural networks for nonlinear feature representation for samples with orientation changes. To be specific, our framework, \emph{i.e.,} OCNs, simultaneously considers finding effective oriented filter bank and learning high-level features from inputs, whereas these existing methods do not embrace the effectiveness of feature learning with orientation changes. We propose a novel method for obtaining effective LGFs and the corresponding coefficients in a matrix factorization framework. Since the sparsity of the coefficients can be increased by the propagation of low-rank structure, we propose an effective integration strategy for the local and global structure, which correspond to sparse coding and low-rank representation, leading to effective LGFs. Then, it is possible to determine responses of any arbitary oriented filters without explicitly applying those filters. We believe that LGFs is complementary with the existing deep neural networks methods since it is able to incorporate the merits of exiting methods into DCNNs. To the best of our knowledge, this is the first attempt to integrate LGFs and DCNNs. Our method is also significantly from \cite{S18}, which just modulates the convolutional filter with the empirically selected Gabor filter bank.

The main contributions in this paper are summarized as follows:
\begin{enumerate}
	\item We define LGFs and the corresponding coefficients in a matrix factorization framework. To achieve effective LGFs, we promote the sparsity of coefficients by propagating the low-rank structure.
	
	\item The proposed LGFs is formulated as an optimization problem with a well-defined objective function. we then design an effectively iterative algorithm to seek for the solution of the constrained optimization problem.
	
	\item We propose OCNs by modulating the basic element of DCNNs  based on LGFs with existing deep learning architecture, which considers LGFs and deep neural networks simultaneously, enforcing the network compact while still obtaining satisfied feature representation ability with orientation changes.
\end{enumerate}

\section{Landmark Gabor Filters}
In this section, we introduce our LGFs for manipulating the convolution filter in DCNNs. The basic purpose of our method is presenting a way for finding effective oriented filter bank, which can employ a limited amount of oriented filters for the accuracy.

We first calculate $ n $ rotated versions of the Gabor filters $ G(u,v) $ based on Eq. (2) and then re-arrange each version into a column vector and then build the matrix  $ X \in R^{m \times n} $. Based on matrix factorization, we propose to find matrices $ U \in R^{m \times p} $ and  $ V \in R^{p \times n} $ whose product can approximate $ X $ via $ \min_{U,V}||X-UV||_{F}^{2} $. To capture the global structure for the set of oriented filters, we put a constraint of low-rank on the LGFs and coefficients for a low-rank representation $ U $ and $ V $. Based on $ V $, we propose to find matrices $ Y \in R^{p \times q} $ and $ Z \in R^{q \times n} $ whose product can best approximate $ V $ via  the constraint $ V=YZ$ and enforce a sparsity constraint on $ Z $ for a sparse representation $ Z $.  It is formulated as follows
\begin{equation}
\begin{split}
\begin{aligned}[alignment]
\min_{U,V,Y,Z} &||X-UV||_{F}^{2}+\lambda rank(U)+\mu rank(V)\\
&+\gamma ||Z||_{0}\;\;s.t. \;V=YZ,
\end{aligned}
\end{split}
\end{equation}
where $ \lambda >0, \;\mu>0, \;\gamma >0 $ are the tradeoff parameters. Note that the objective function in (7) is NP-hard for the rank function $ rank(\;.\;) $ and the $ l_{0} $-norm $||\;.\;||_{0} $ on their minimizations. As is known, the nuclear norm $||\;.\;||_{*} $ is widely adopted and has been proved to be the convex envelop of $ rank(\;.\;) $. Meanwhile, the $ l_{1} $-norm $ ||\; .\;||_{1} $ has been demonstrated as an effective approximations of $ l_{0} $-norm $ ||\;.\;||_{0} $. Then we can obtain:
\begin{equation}
\begin{split}
\begin{aligned}[alignment]
\min_{U,V,Y,Z} &||X-UV||_{F}^{2}+\lambda ||U||_{*}+\mu ||V||_{*}+\gamma ||Z||_{1} \\ 
&s.t. \;V=YZ,
\end{aligned}
\end{split}
\end{equation}
where the nuclear norm $ ||U||_{*} =\sum_{i}|s_{i}|$. $ s_{i} $ is the singular value of the matrix $ U $ and likewise $||V||_{*} $. Obviously, with the lpropagation, the sparse representation $ Z $ can encode with both the global and local structure. Since $ U \in R^{m \times p}, Y \in R^{p \times q} $ and $ UY \in R^{m \times q} ,Z \in R^{q \times n}$, each column of $ UY $ can be seen as a LGF capturing high-level features of the original oriented filter. To better deal with the noise, \emph{i.e.,} Gaussian noise, we relax the constraint of (8). This yields the optimization problem as follows 
\begin{equation}
\begin{split}
\begin{aligned}[alignment]
\min_{U,V,Y,Z} &||X-UV||_{F}^{2}+\lambda ||U||_{*}+\mu ||V||_{*}+\gamma ||Z||_{1} \\ 
&+\rho||V-YZ||_{F}^{2},
\end{aligned}
\end{split}
\end{equation}
where least squares, nuclear-minimizations and $ l_{1} $-minimizations are involved in (9). The problem to be solved in (9) is non-convex regarding joint $ (U,V,Y,Z) $. We then employ the efficient iterative algorithm \cite{CL18} to seek for the solution of the optimization problem in (9). We solve the nuclear-minimization and $ l_{1} $-minimization in (9) by the Singular Value Shrinkage Thresholding algorithm (SVT) \cite{A.B09} and the Iterative Shrinkage Thresholding algorithm (IST) \cite{J.C10}, respectively, which are defined in the following.

\textit{Lemma 1} (\textbf{Iterative Shrinkage Thresholding} \cite{J.C10}) \textit{Given a matrix M and a positive number} $ \tau $, \textit{the following minimization}
\begin{equation}
\min_{X} \tau ||X|_{1}+1/2||X-M||^{2}_{F}
\label{XX}
\end{equation}
\textit{has a global optimal solution in the closed form}
\begin{equation}
X=sign(M)max(0,|M|-\tau).
\end{equation}

\textit{Lemma 2} (\textbf{Singular Value Shrinkage Thresholding}\cite{A.B09}) \textit{Given a matrix }$ M=USV^{T} $ \textit{and a positive number }$ \tau $, \textit{the following minimization}
\begin{equation}
\min_{X} \tau ||X|_{*}+1/2||X-M||^{2}_{F}
\label{XX}
\end{equation} 
\textit{has a  global optimal solution in the closed form}
\begin{equation}
X=Usign(S)max(0,|S|-\tau)V^{T}.
\end{equation}
Based on the Alternating Direction Minimization (ADM) strategy, we divide the objective function into the following subproblems:

(1) $Y$-\textit{Subproblem}: The $Y$-subproblem is shown as
\begin{equation}
\begin{split}
\begin{aligned}[alignment]
Y^{*}=\arg\min\limits_{Y}\;\rho||V-YZ||_{F}^{2},
\end{aligned}
\end{split}
\end{equation}
Taking the derivative with respect to $ U $ and then enforcing it to be zero, we obtain the closed-form solution as:
\begin{equation}
Y^{*}=V^{T}Z(Z^{T}Z)^{-1}.
\end{equation}

(2) Updating $ U $: The LGFs $ U $ is updated by
 \begin{equation}
\begin{split}
\begin{aligned}[alignment]
U^{*}=\arg\min\limits_{U}\;||X-UV||_{F}^{2}+\lambda ||U||_{*},
\end{aligned}
\end{split}
\end{equation}

(3) Updating $ V $: The coefficient $ V $ with respect to $ U $ is updated by
\begin{equation}
\begin{split}
\begin{aligned}[alignment]
V^{*}=\arg\min\limits_{V}\;||X-UV||_{F}^{2}+\lambda ||V||_{*}+\rho||V-YZ||_{F}^{2},
\end{aligned}
\end{split}
\end{equation}

(4) Updating $ Z $: The coefficient $ Z $ regarding $ Y $ is updated by
\begin{equation}
\begin{split}
\begin{aligned}[alignment]
Z^{*}=\arg\min\limits_{Z}\;\gamma ||Z||_{1}+\rho||V-YZ||_{F}^{2},
\end{aligned}
\end{split}
\end{equation}

We solve the subproblems corresponding to $ U $ and $ V $ by the SVT algorithm. IST can be adopted to deal with the subproblem corresponding to $ Z $. For clarification, we summarize the optimization procedure in Algorithm 1.
\begin{algorithm}[!ht]
	\caption{Algorithm of the proposed LGFs}
	\SetKwInOut{Input}{Input}\SetKwInOut{Output}{Output}\SetKwInOut{Initialize}{Initialize}
	\Input {Set of oriented filters $ X=\left[ x_{1},x_{2},...,x_{n}\right]  \in R^{m \times n} $, regularization parameters $ \rho, \;\lambda, \;\mu, \;\gamma$.}
	\Output {Basis matrix $ U,Y $ and coefficient matrix $ V,\;Z $.}
	\Initialize {Randomly initialize $ U,\;V,\;Y,\;Z $ and set $ i=1 $.}
	\Repeat{$ convergence $}{
			Fix $ U,\;V$ and $Z $, update $ Y $ by updating rule (15);\\
			Fix $ V,\;Y$ and $Z$, update $ U $ according to Eq. (16);\\
		 	Fix $ U,\;Y$ and $Z $, update $ V $ according to Eq. (17);\\
		    Fix $ U,\;V$ and $Y $, update $ Z $ according to Eq. (18);\\
			$ i=i+1 $.
		}
	\end{algorithm}

\textbf{Convergence Analysis: } The problem in (9) is not convex regarding the joint variables $ \left( U,V,Y,Z\right)  $. The objective function in (9) is convex regarding each single variable in $ \left( U,V,Y,Z\right)$ since only $ l_{1} $-norm, Frobenius-norm and nuclear-norm are involved and all of them are convex functions.

In the following, we theoretically show that the optimal solution $ \left( V,Z\right)  $ to (9) is the final clustering result. Inspired by \cite{ZL18} that we can explicitly pursue the block-diagonal structure in the matrix factorization framework, we give a theoretical proof of the optimal solution $ \left( V,Z\right)$ is block diagonal. Meanwhile, we also show why $ Z $ is more proper as the coefficient.

\textbf{Theorem 1}. \textit{Let} $X \in R^{m\times n}$ \textit{be a set of oriented filters whose columns are sampled from a union of subspaces which are independent. Assume that} $ X $ \textit{has been sorted based on the subspaces, i.e.}, $ X=\left[ x_{1},x_{2},...x_{n}\right] \Gamma $, \textit{where} $ \Gamma $ \textit{is an permutation matrix specifying subspaces of original oriented filters. Then there is the optimal solution } $\left( V,Z\right)$ \textit{to } (9) \textit{with} $ V $ \textit{and }$ Z $ \textit{being both}  \textit{block diagonal, i.e., there are just connections within subspace and no connection between subspaces}.

\textbf{Proof}. Let $\left( V,Z\right)$ be the optimal solution to (9). Note that different subspaces distinguish each other with different oriented basis filters, \emph{i.e.,} LGFs, we can use the oriented basis filter to represent the subspace for clarity. Since $ V $ is a solution of the nuclear norm problem in (9), then $ v_{i} $ is a vector of minimum nuclear norm satisfying $ x_{i}=Uv_{i} $. Let us decompose $ v_{i} $ as $ v_{i}=v_{i}^{'}+ h_{i}$, where $ v_{i} ^{'}$ denotes the coefficient recovered from the true basis filter of $ x_{i} $ and $ h_{i} $ denotes the coefficient recovered from the other basis filter. Now we just need to prove $ h_{i}=0 $. Then, for any oriented filter $ x_{i} $, since $ v_{i}=v_{i}^{'}+ h_{i} $, we have $x_{i}=Uv_{i}=U(v_{i}^{'}+h_{i})=Uv_{i}^{'}+Uh_{i}$. Moreover, since $x_{i} \in U_{i}$ , $Uv_{i} \in U_{i}$ and $ Uh_{i} \notin U_{i}$, we can obtain $ Uh_{i}=0 $. Based on the fact that $ v_{i}^{'} $ and $ h_{i} $ have disjoint subset of indices, if $ h_{i} \notin 0 $ is assumed, we can obtain $ ||v_{i}^{'}||_{*}<||v_{i}^{'}+h_{i}||_{*}=||v_{i}||_{*} $, which contradicts the optimality of $ v_{i} $. Then, $ h_{i}=0 $ and $ V $ is block diagonal.

Moreover, based on $ V $ , if $ Z $ is the solution of the $ l_{1} $-norm problem in (9), then $ z_{i} $ is a vector of the minimum $ l_{1} $-norm satisfying $ x_{i}=Wz_{i} $. Let us decompose $ z_{i} $ as $ z_{i}=z_{i}^{'}+o_{i} $, where $ z _{i}^{'}$ denotes the coefficient recovered from the true basis filter of $ x_{i} $ and $ o_{i} $ denotes the coefficient recovered from the other basis filter. As above, we just need to prove $ o_{i}=0 $. Since $ z_{i}=z_{i}^{'}+o_{i} $, then $ x_{i}=Wz_{i}=W(z_{i}^{'}+o_{i} ) =Wz_{i}+Wo_{i}$ can be obtained. Besides, since $ x_{i} \in W_{i}, Wz_{i} \in W_{i} $ and $ Wo_{i} \notin W_{i} $, we can obtain $ Wo_{i}=0 $. Since $ z_{i}^{'} $ and $ o_{i} $ have the disjoint subset of indices, if $ o_{i} \notin 0 $ is assumed, we can obtain $ ||z_{i}^{'}||_{1}<||z_{i}^{'}+o_{i}||_{1}= ||z_{i}||_{1}$, which contradicts the optimality of $ z_{i} $. Therefore, we can obtain $ o_{i}=0 $ and $ Z $ is block diagonal, which concludes the proof.

Since the solution $ V$ may not be unique, which has been proved in \cite{2013Robust1}, Theorem 1 is not able to ensure that any optimal $ V $ is block diagonal with an unknown permutation $ \Gamma $. However, according to Theorem 1, any optimal $ Z $ can be guaranteed to be block diagonal under an unknown permutation $ \Gamma $. This also explains why we choose $Z$ instead of $V$ as our coefficient. Moreover, we can observe that $ Z $ incorporates both the global and local structure of original oriented filters. However, $ V $ just encodes the global structure. Then, $ UY\in R^{m\times q} $ with the coefficient $ Z\in R^{q\times n} $ is used as our LGFs.
\section{Orientation Convolutional Networks}
In order to incorporate the orientation and scale information into OCNs, we modulate the convolutional filters in standard CNNs with LGFs. Then, the corresponding convolutional features in OCNs can be enhanced. Unlike standard DCNNs, which express the dimensions of the weight per layer as $ C_{out}\times C_{in}\times W \times W$, OCNs denote it as  $ C_{out}\times C_{in}\times N^{'}\times W \times W$, where $ W\times W $ is the size for the filter, $ N^{'} $ is the channel, $ C_{out} $ and $ C_{in} $ refer to the channel for the corresponding feature maps. Here, $ N^{'} $ is set by the number of orientations $ U $. For the given scale $ v $, we define the modulated procedure using LGFs in DCNNs as:
\begin{equation}
\begin{split}
\begin{aligned}[alignment]
C_{i,u}^{v}=C_{i,o}\odot LG(u,v),
\end{aligned}
\end{split}
\end{equation}
where $C_{i,o}$ denotes the learned filter, and $ \odot $ is a operation for the element-by-element product between each 2D filter of $C_{i,o}$ and $LG(j,v)$. $ C_{i,j}^{v} $ is the modulated one. The obtained modulated convolution filters are defined as:
\begin{equation}
\begin{split}
\begin{aligned}[alignment]
C_{i}^{v}=\left(C_{i,1}^{v}, C_{i,2}^{v},...,C_{i,U}^{v} \right). 
\end{aligned}
\end{split}
\end{equation}

For each scale, we can achieve enhanced features in this way. In the following, we omit the scale $ v $ to simplify the description process.

The obtained modulated convolution filters are basic element of OCNs, which can generate feature maps encoding the orientation and scale information. The output feature map $\widehat{F}$ in OCNs is defined as:
\begin{equation}
\begin{split}
\begin{aligned}[alignment]
\widehat{F}_{i,k}=\sum\limits_{n=1}^{N^{'}}F^{n}\otimes C_{i,u=k}^{n}\label{XX}
\end{aligned}
\end{split}
\end{equation}
where $ N^{'} $ refers to the channel, $ F $ is the input feature map, $ n $ is the \textit{n}-th channel of $F$. $ C_{i,u} $. $ \widehat{F}_{i,k} $ refers to the \textit{k}-th response of $F_{i}$. 

In OCNs, the weights contained in the forward convolution process are the modulated convolution filters. However, the weights needed to be saved are just the learned filters. Then, we update the learned filter $C_{i,o}$ in the back-propagation (BP) process. By summing up the gradient of all filters in the modulated convolution filters, we can obtain:
\begin{equation}
\begin{split}
\begin{aligned}[alignment]
&C_{i,o}=C_{i,o}-\eta\upsilon, \\
&\upsilon=\frac{\partial Loss}{\partial C_{i,o}}=\sum_{u=1}^{U}\frac{\partial Loss}{\partial C_{i,u}}\odot LG(u,v),\label{XX}
\end{aligned}
\end{split}
\end{equation}
where $ Loss $ refers to the loss function and $ \eta $ is the learning rate in the BP process. It can be found that the BP process in our method is the same as the standard BP process and  we just need to update the learned convolution filters $ C_{i,o} $, resulting in a more compact and efficient model.
\section{Experiments and Analysis}
In this section, we implement OCNs based on the conventional CNNs and the ResNet, which are listed in Fig. 2. We evaluate the proposed OCNs on MNIST digital recognition dataset, the rotated version MNIST-rot dataset, CIFAR-10 dataset, CIFAR-100 dataset. NVIDIA GeForce GTX 1080Ti is used in our experiment. We first investigate how parameters in (9) influence the performance and show the selection of parameters in the experiment.

\subsection{Parameters Selection}
To investigate how parameters $ (\rho, \mu, \lambda, \gamma)$ in (9) influence the performance, we use $D=\frac{||\hat{X}||_{F}^{2}}{||X||_{F}^{2}}  $ as a measurement of the approximation degree of original oriented filters. We seek for $ (\rho, \mu, \lambda, \gamma)$ corresponding to the maximal degree of approximation. Then the problem becomes seeking for the maximal $D$. 

For clarity, we employ the original set of oriented filters in the following. The set of oriented filters are based on the Gabor filter and generated as: the central frequency $ F=16$ and the orientation sample interval of $ 5^{o} $ are used. For Gabor filters with central frequency 16, 5 LGFs are used. Since the initial result shows that parameters should be set to small values, we choose in $\left\lbrace 0.001,0.005,0.01,0.05,0.1,0.5\right\rbrace$ for investigations. To reduce the computation complexity, we fist set $\rho= \mu= \gamma$ and find the optimal $ \lambda $. Then we find the optimal $\rho, \mu, \gamma$ with the fixed optimal $\lambda $. Likewise, we set $ \rho= \gamma$ and find the optimal $\mu$. Then we find the optimal $\rho, \gamma$ with the fixed optimal $\mu$. According to Fig. 1, we can set $ \rho=0.05,\; \mu=0.05,\; \gamma=0.05,\; \lambda=0.05 $ for the four parameters for simplicity.

\begin{figure} [!htbp]
	\centering    
	
	\subfigure[] 
	{
		\begin{minipage}[t]{0.4\textwidth}
			\centering          
			\includegraphics[width=\textwidth]{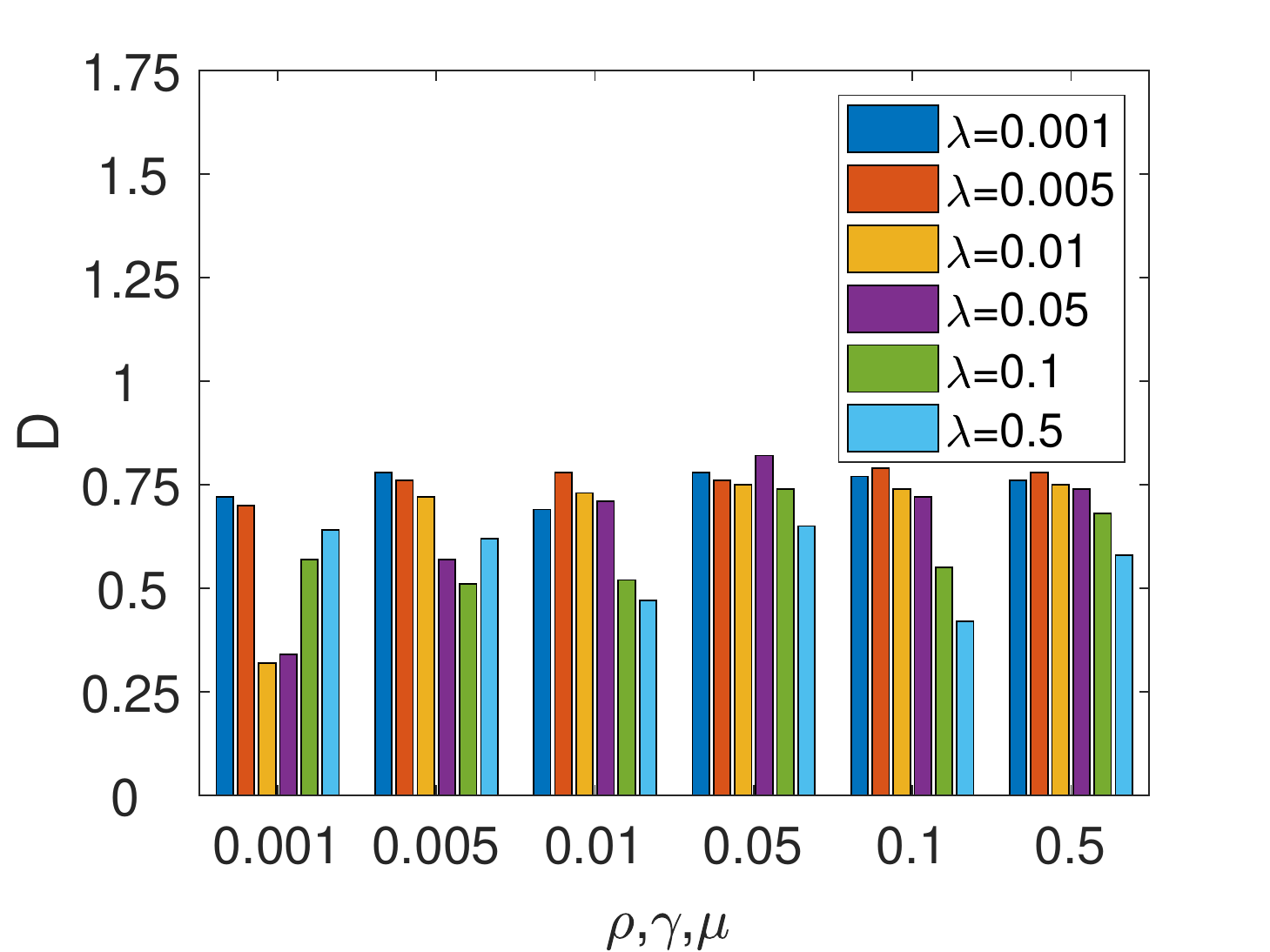}
		\end{minipage}%
	}
	
	\subfigure[] 
	{
		\begin{minipage}[t]{0.4\textwidth}
			\centering      
			\includegraphics[width=\textwidth]{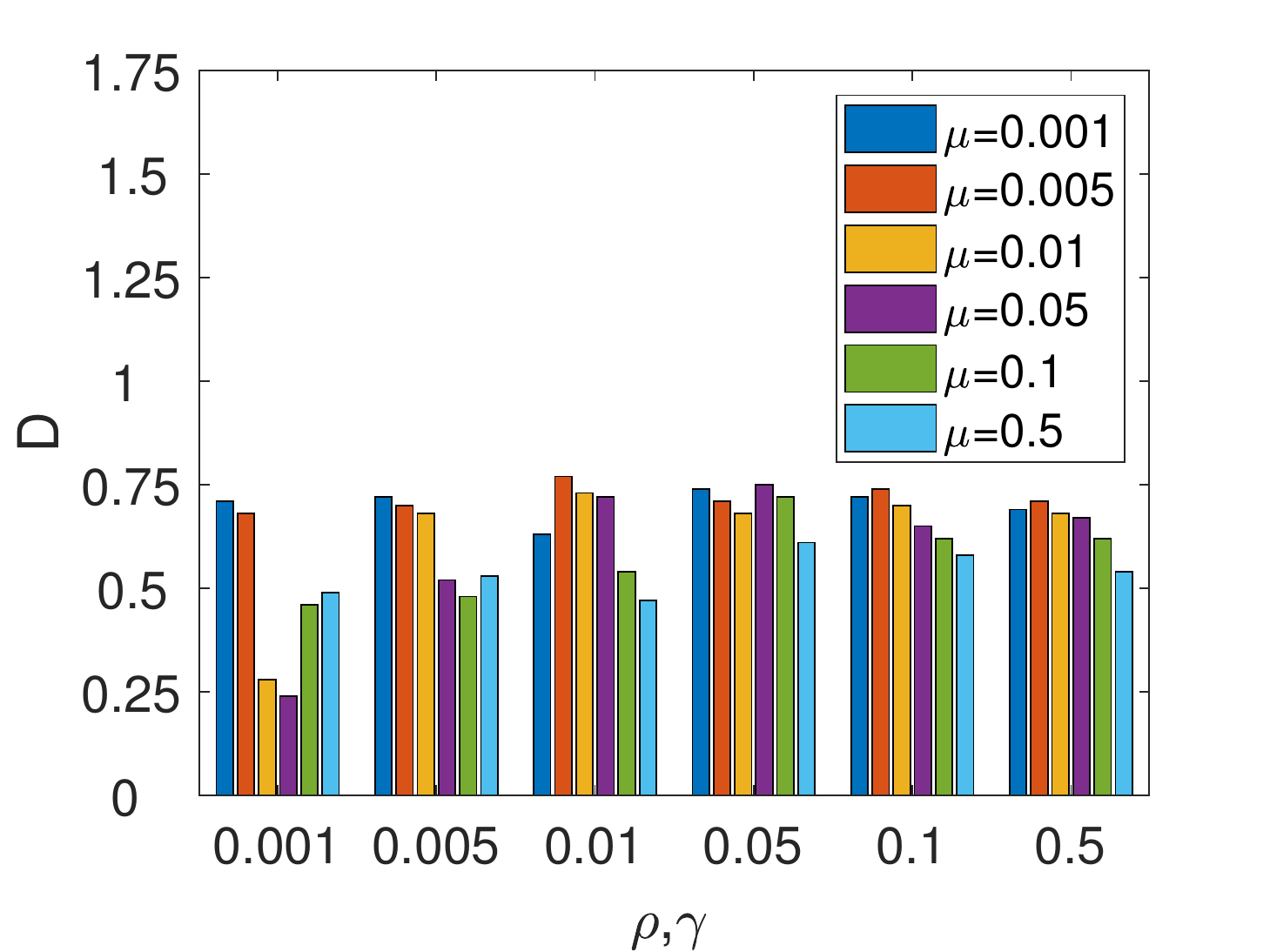}
		\end{minipage}
	}
		\subfigure[] 
	{
		\begin{minipage}[t]{0.4\textwidth}
			\centering      
			\includegraphics[width=\textwidth]{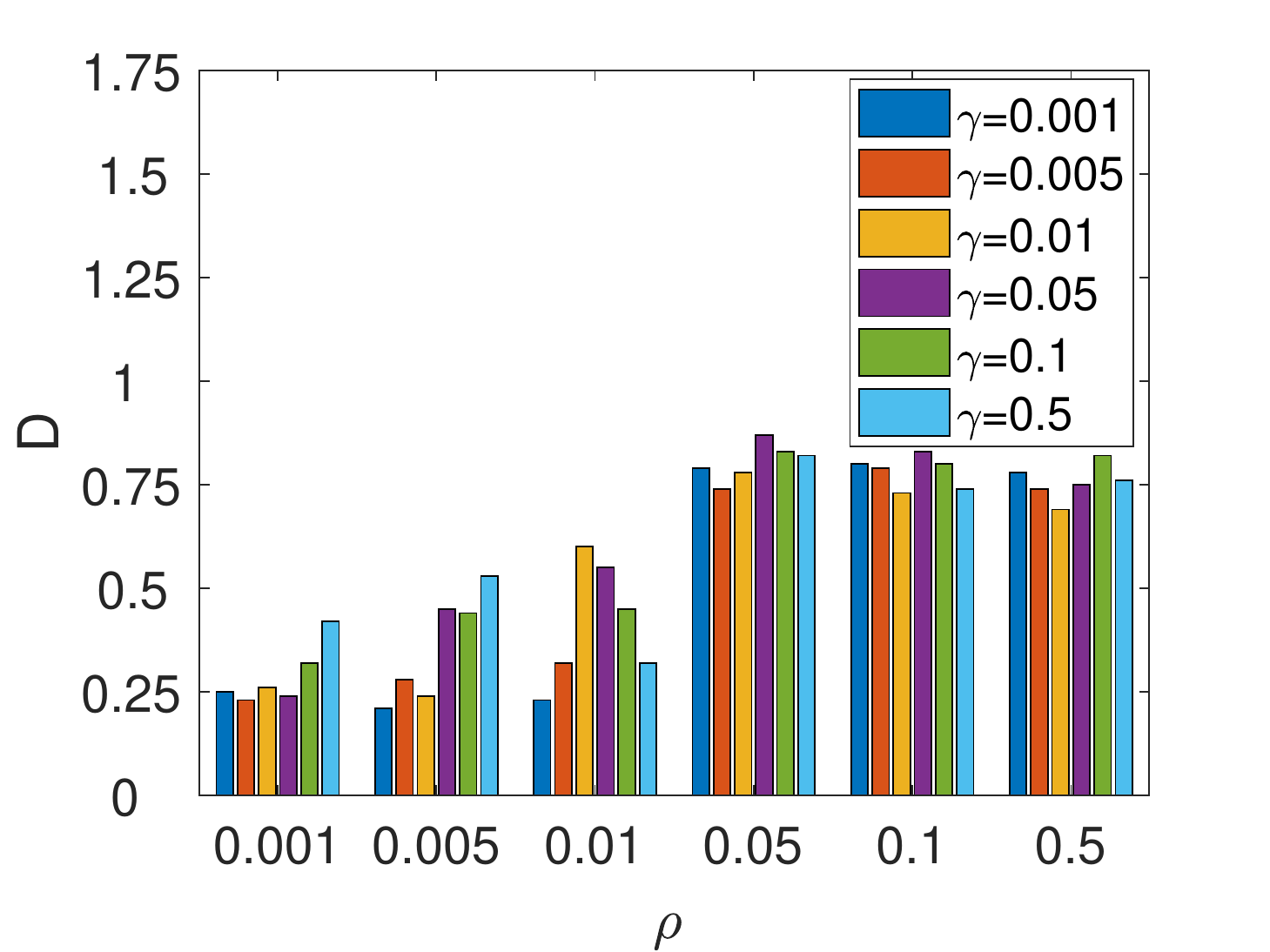}
		\end{minipage}
	}
	\caption{Degree of approximation of oriented filters: (a) $ \lambda $ vs $(\rho, \gamma, \mu)  $; \;(b) $ \mu $ vs $ (\rho, \gamma) $; (c) $ \gamma $ vs $ \rho $.}
	\label{fig1}  
\end{figure}
\begin{figure*}[ht]
	\vskip 0.2in
	\begin{center}
		\centerline{\includegraphics[width=\textwidth]{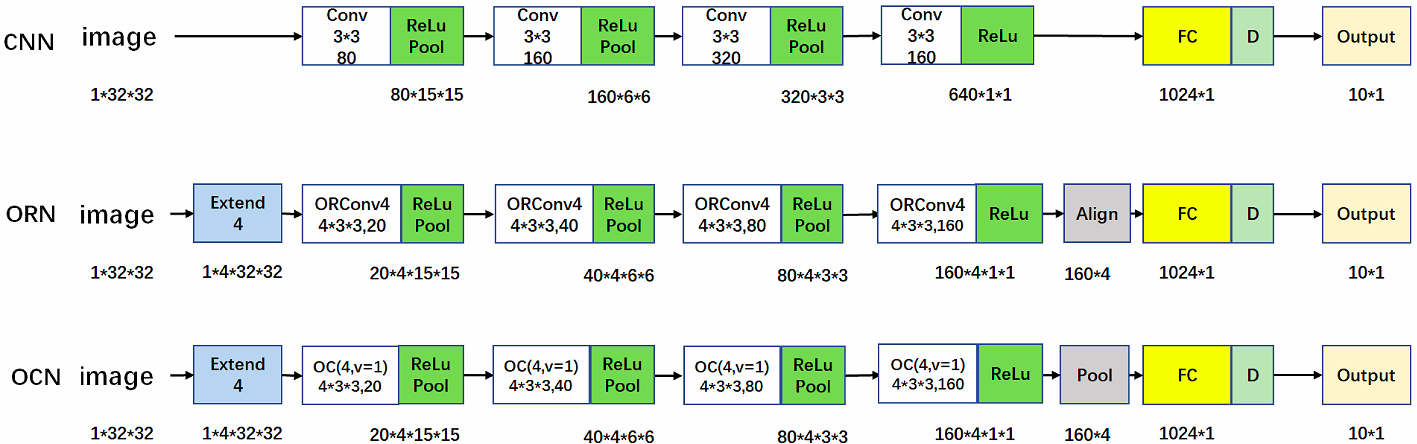}}
		\caption{Network structure of CNN, ORN and OCNs.}
		\label{icml-historical}
	\end{center}
	\vskip -0.2in
\end{figure*}
\begin{table}[t]
	\caption{Error rate $ (\%) $ on MNIST when V=1.}
	\label{sample-table}
	\vskip 0.15in
	\begin{center}
		\begin{small}
			\begin{sc}
				\begin{tabular}{|c|c|c|}
					\hline
					Size&10-20-40-80 &20-40-80-160\\
					\hline
					\hline
					$ 5\times5 $ &0.58 &0.59\\	
					\hline
					$ 3\times3 $ &0.62 &0.60\\				
					\hline
				\end{tabular}
			\end{sc}
		\end{small}
	\end{center}
	\vskip -0.1in
\end{table}

\begin{table}[t]
	\caption{Error rate $ (\%) $ on the MNIST when V=4.}
	\label{sample-table}
	\vskip 0.15in
	\begin{center}
		\begin{small}
			\begin{sc}
				\begin{tabular}{|c|c|c|}
					\hline
					Size&10-20-40-80 &20-40-80-160\\
					\hline
					\hline
					$ 5\times5 $ &0.57 &0.51\\	
					\hline
					$ 3\times3 $ &0.62 &0.59\\			
					\hline
				\end{tabular}
			\end{sc}
		\end{small}
	\end{center}
	\vskip -0.1in
\end{table}

\begin{table}[t]
	\caption{Error rate $ (\%) $ on the MNIST of different orientations.}
	\label{sample-table}
	\vskip 0.15in
	\begin{center}
		\begin{small}
			\begin{sc}
				\begin{tabular}{|c|c|c|c|c|c|c|}
					\hline
					Size&2 &3 &4 &5 &6 &7\\
					\hline
					\hline
					$ 5\times5 $ &0.56 &0.54 &0.51 &0.43 &0.44 &0.48\\	
					\hline
					$ 3\times3 $ &0.623 &0.58 &0.59 &0.54 &0.54 &0.59\\				
					\hline
				\end{tabular}
			\end{sc}
		\end{small}
	\end{center}
	\vskip -0.1in
\end{table}
\begin{table*}[h]
	\caption{Performace comparison on the MNIST.}
	\centering
	\begin{tabular}{cccccc}
		\hline
		\multicolumn{4}{c}{Method} & \multicolumn{2}{c}{Error$ (\%) $}\\
		\hline
		&network stage kernel&params (M)&times (s)&MNIST&MNIST-rot\\
		\hline
		Baseline CNN&80-160-320-640&3.08&6.50&0.73&2.82\\
		STN&80-160-320-640&3.20&7.33&0.61&2.52\\
		TIpooling$ (\times 8) $& (80-160-320-640)$ (\times 8) $&3.08&50.21&0.97&-\\
		ORN4(ORAlign)&10-20-40-80&0.49&9.21&0.57&1.69\\
		ORN8(ORAlign)&10-20-40-80&0.96&16.01&0.59&1.42\\
		ORN4(ORPooling)&10-20-40-80&0.25&4.60&0.59&1.84\\
		ORN8(ORPooling)&10-20-40-80&0.39&6.56&0.66&1.37\\
		GCN4{$ (3\times 3)$}&10-20-40-80&0.25&3.45&0.63&1.45\\
		GCN4{$ (3\times 3)$}&20-40-80-160&0.78&6.67&0.56&1.28\\
		GCN4{$ (5\times 5)$}&10-20-40-80&0.51&10.45&0.49&1.26\\
		GCN4{$ (5\times 5)$}&20-40-80-160&1.86&23.85&0.48&1.10\\
		GCN4{$ (7\times 7)$}&10-20-40-80&0.92&10.80&0.46&1.33\\
		OCN4{$ (7\times 7)$}&20-40-80-160&3.17&25.17&0.42&1.20\\
		OCN4{$ (3\times 3)$}&10-20-40-80&0.25&3.43&0.62&0.81\\
		OCN4{$ (3\times 3)$}&20-40-80-160&0.78&6.51&0.59&0.70\\
		OCN4{$ (5\times 5)$}&10-20-40-80&0.51&9.47&0.57&\textbf{0.55}\\
		OCN4{$ (5\times 5)$}&20-40-80-160&1.86&22.79&0.51&0.65\\
		OCN4{$ (7\times 7)$}&10-20-40-80&0.92&12.70&0.41&0.56\\
		OCN4{$ (7\times 7)$}&20-40-80-160&3.47&29.02&\textbf{0.40}&0.55\\
		\hline					
	\end{tabular}
	\label{tab:Margin_settings}
\end{table*}
\begin{figure}[ht]
	\vskip 0.2in
	\begin{center}
		\centerline{\includegraphics[width=\columnwidth]{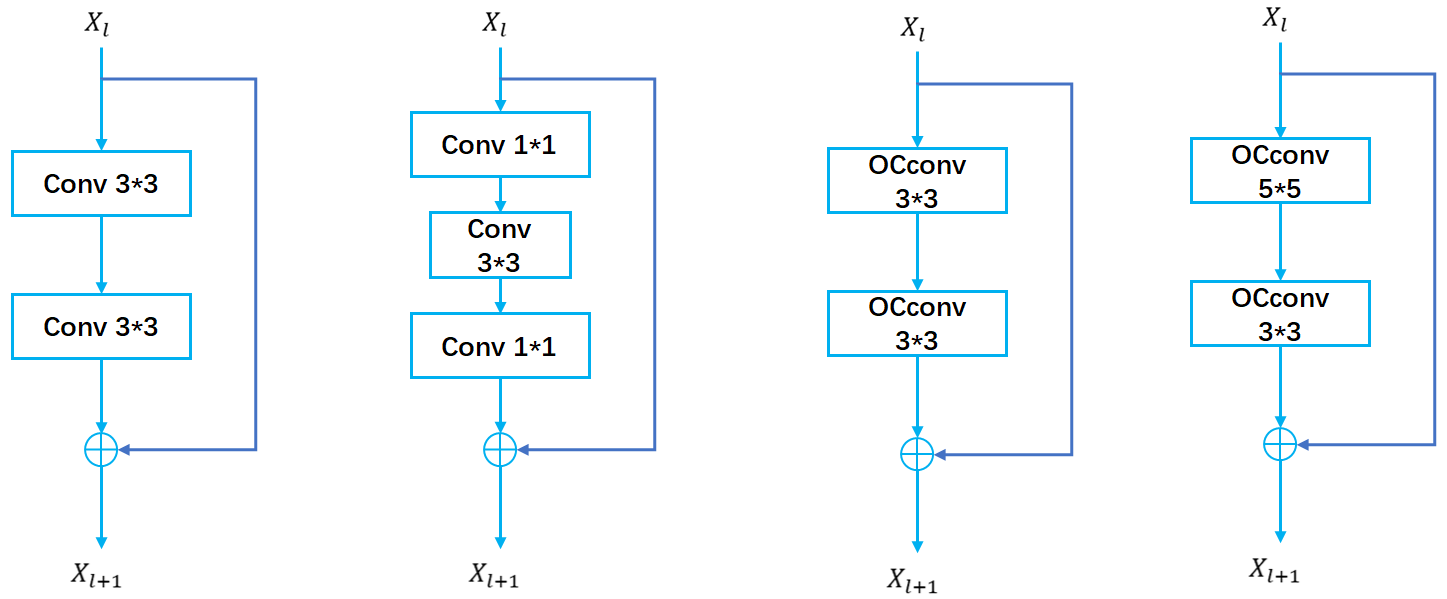}}
		\caption{The residual block. The former two are used for ResNet. The third one is small kernel for OCNs. The last one in large kernel for OCNs.}
		\label{icml-historical}
	\end{center}
	\vskip -0.2in
\end{figure}

\begin{figure}[ht]
	\vskip 0.2in
	\begin{center}
		\centerline{\includegraphics[width=\columnwidth]{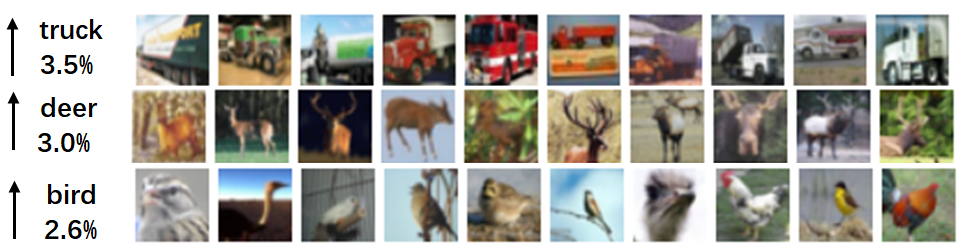}}
		\caption{Sample images which contain rotated objects/parts falsely classified by ResNet but correctly classified by the proposed OCNs in CIFAR-10.}
		\label{icml-historical}
	\end{center}
	\vskip -0.2in
\end{figure}

\begin{table*}[h]
	\caption{Performace comparison on the CIFAR-10 and the CIFAR-100.}
	\centering
	\begin{tabular}{ccccc}
		\hline
		\multicolumn{3}{c}{Method}& Error on CIFAR-10$ (\%) $& Error on CIFAR-100$ (\%) $\\
		\hline
		\multicolumn{3}{c}{NIN}&8.81&35.67\\
		\multicolumn{3}{c}{VGG}&6.32&28.49\\
		\hline
		&network stage kernel&params (M)&&\\
		\hline
		ResNet-110&16-16-32-64&1.7&6.43&25.16\\
		ResNet-1202&16-16-32-64&10.2&7.83&27.82\\
		GCN2-40&16-32-64-128&4.5&4.95&24.23\\
		GCN2-110&12-12-24-45&1.7/3.3&6.34/5.62&-\\
		GCN2-110&16-16-32-64&3.4/6.5&5.65/4.96&26.14/25.3\\
		GCN4-110&8-8-16-32&1.7&6.19&-\\
		GCN4-40&16-16-32-64&2.2&5.34&25.65\\
		OCN2-110&12-12-24-45&1.7/3.3&6.32/5.34&-\\
		OCN2-110&16-16-32-64&3.4/6.5&5.65/5.95&26.35/25.5\\
		OCN4-110&8-8-16-32&1.7&6.07&-\\
		OCN2-40&16-32-64-128&4.5&4.83&22.34\\
		OCN4-40&16-16-32-64&2.2&5.21&24.16\\
		WRN-40&64-64-128-256&8.9&4.53&21.18\\
		WRN-28&160-160-320-640&36.5&4.00&\textbf{19.25}\\
		GCN2-40&16-64-128-256&17.9&4.41&20.73\\
		GCN4-40&16-32-64-128&8.9&4.65&21.75\\
		GCN3-40&64-64-128-256&17.6&3.88&20.13\\
		OCN2-40&16-64-128-256&17.9&4.21&20.76\\
		OCN4-40&16-32-64-128&8.9&4.75&22.04\\
		OCN3-40&64-64-128-256&17.6&\textbf{3.76}&19.45\\
		\hline					
	\end{tabular}
	\label{tab:Margin_settings}
\end{table*}

\subsection{MNIST and MNIST-rot}
MNIST dataset has 60,000 data points used for training and 10,000 data points used for testing. For our method, we set the batch size as 128 and the weight decay as 0.00005. Meanwhile, the initial learning rate is set as 0.001 and the learning rate is reduced to half per 10 epochs. In the following, we show the performance of our method on the testing set based on the average 5 runs after 50 epochs. For baselines, we tune their parameters to achieve the best results. For comparision, we use different algorithms including STN\cite{Jaderberg15}, TI-Pooling\cite{Laptev16}, GCNs\cite{S18} and ORNs\cite{Zhou17}. For clarity, in OCNs, we use OConv (OC) as the name of the enhanced convolution filter which is modulated with LGFs generated by our method.

Note that network structures of conventional CNNs, ORNs and ours are shown in Fig. 2. It can be seen that Max-pooling and ReLu are used after convolutional layers for all network structures. Meanwhile, dropout layer is used for avoiding over-fitting. For comparison with the other CNNs, the size of similar model is guaranteed by reducing the width of the convolutional kernel \cite{Zagoruyko16} as done in ORNs. We first test several scales for different OCNs layers, \emph{i.e.,} $ V=1, V=4 $. According to Tables \uppercase\expandafter{\romannumeral1}-\uppercase\expandafter{\romannumeral2}, we can observe that the performance of $ V=4 $ is better than that of $ V=1 $ in terms of error rate. The number of orientations is also evaluated in Table \uppercase\expandafter{\romannumeral3} when $ V=4$, which shows that OCNs can achieve satisfied performance when 3 to 7 orientations are used. It can be found that enough orientation channels are needed to capture the oriented information.

We also compare the proposed OCNs with the baseline methods on MNIST and MNIST-rot dataset. As is shown in Table \uppercase\expandafter{\romannumeral4}, the last two columns show the performance comparison in terms of the error rate. When compared with the baseline conventional CNNs, OCNs show better performance with $ 3 \times 3 $ kernel size but using fewer parameters of CNNs. Besides, it can be observed that the testing error of OCNs with $ 5\times 5 $ and $ 7 \times 7 $ kernel size is $ 0.55 \%$ on MNIST-rot and $ 0.40\% $ on MNIST dataset. It is due to the fact that larger kernel sizes have more oriented information. Besides, we list the corresponding computation time of compared methods, which demonstrate that OCNs are more effective than other state-of-art baseline models. Table \uppercase\expandafter{\romannumeral4} also shows that a larger OCNs can indeed produce better performance, which validates that oriented filters approximated by our method help to achieve the robustness to rotation variations.
\begin{table*}[t]
	\caption{Performance comparison on the SVHN without using the additional training set.}
	\label{sample-table}
	\vskip 0.15in
	\begin{center}
		\begin{small}
			\begin{sc}
				\begin{tabular}{|c|c|c|c|c|c|c|c|c|}
					\hline
					Method&VGG&ResNet-110&ORN4-40&ORN4-28&GCN4-40&GCN4-28&OCN4-40&OCN4-28\\
					\hline
					\hline
					Params(M)&20.3&1.7&2.2&1.4&2.2&1.4&2.2&1.4\\
					\hline
					Accuracy$ (\%) $ &95.66&95.8&96.35&96.19&96.9&96.86&97.1&96.98\\				
					\hline
				\end{tabular}
			\end{sc}
		\end{small}
	\end{center}
	\vskip -0.1in
\end{table*}
\subsection{CIFAR10 and CIFAR100}
The CIFAR dataset which contains CIFAR-10 and CIFAR-100 is used for classification issue of the natural image. It has 50,000 images used for training and 10,000 images used for testing. CIFAR-10 consists of 10 classes with 6,000 images contained per class and CIFAR-100 has 100 classes with 600 images contained per class. The size of each colored image is $ 32 \times 32 $.

The network structure used is based on ResNet \cite{He06}.  For clarity, with the enhanced convolution filter as above, we use OCN-ResNets as the name of the network. For baselines, we tune their related parameters to achieve the best results. Table \uppercase\expandafter{\romannumeral5} depicts results of different algorithms such as NIN \cite{Lin14} and ResNet. On CIFAR-10, Table \uppercase\expandafter{\romannumeral4} indicates that OCN-ResNets indeed improve the performance both in the different numbers of parameters and kernel size compared with the baseline ResNet. Furthermore, the result compared with the Wide Residue Network (WRN)\cite{Zagoruyko16} shows that OCN-ResNets can still obtain a better performance when OCN-ResNets are half the size of WRN, demonstrating its advantage in terms of the model efficiency. Due to the huge parameters of WRN and more categories in CIFAR-100 data set, we can still observe that WRN better performs on CIFAR-100 data set. In addition, according to Fig. 4, we can observe that the top improved categories in CIFAR10 dataset are truck $(3.5\%)$, bird $(2.6\%)$, and deer $(3.0\%) $, indicating significant scale variations exist within category. It demonstrates that OCN-ResNet indeed enhances the capability of dealing with scale variation. 

\subsection{SVHN dataset}
The Street View House Numbers (SVHN) dataset \cite{Netzer11} is an image dataset. It has $ 32\times32 $ images centered around a single character like MNIST. There are 531,131 additional images in this dataset. In the experiments, we do not use the additional images for all models. For such dataset, we implement OCNs based on ResNet. Specifically, the spatial convolutional layers are replaced with the OCconv layer based on modulated convolution filters, resulting in OCN-ResNet. Besides, the bottleneck structure is not adopted due to the fact that $ 1\times 1 $ kernel does not propagate any information of modulated convolution filters. The whole network is divided into 4 stages by ResNet. To ensure that the proposed OCNs method has the similar model size as the compared ResNet, we adjust the depth and width of the network. In our experiment, the hyper-parameters for 40-layer and 28-layer OCN-ResNets are set  the same as ResNet. The network stage is set as 16-16-32-64. In Table \uppercase\expandafter{\romannumeral6}, the performances of different methods are shown. OCN-ResNet has much smaller parameter size than VGG model, yet an improved performance with $ 1.32\% $ is obtained. Without an increased parameter size, the OCN-ResNet shows a better performance than ResNet and GCNs respectively, which demonstrates the effectiveness of OCN-ResNet for real world images.
\section{Conclusion}
This paper studies the problem of obtaining effective oriented filters, \emph{i.e.,} LGFs, and proposes a new deep learning model by combining LGFs and DCNNs for enhanced representations. The main contribution is designing LGFs, as well as improving state-of-the-art DCNNs architectures on the generalization ability towards orientation and scale variations. LGFs can be easily implemented on the exiting architectures and the whole process of designing LGFs is flexible and extensible, which makes it better modulate the standard filter in DCNNs. Experimentally, OCNs significantly outperform the baselines, obtaining the sate-of-the-art performance over benchmarks.

\bibliographystyle{IEEEtran}

\bibliography{IEEEabrv,IEEEexample}

\begin{thebibliography}{10}
\providecommand{\url}[1]{#1}
\csname url@samestyle\endcsname
\providecommand{\newblock}{\relax}
\providecommand{\bibinfo}[2]{#2}
\providecommand{\BIBentrySTDinterwordspacing}{\spaceskip=0pt\relax}
\providecommand{\BIBentryALTinterwordstretchfactor}{4}
\providecommand{\BIBentryALTinterwordspacing}{\spaceskip=\fontdimen2\font plus
\BIBentryALTinterwordstretchfactor\fontdimen3\font minus
  \fontdimen4\font\relax}
\providecommand{\BIBforeignlanguage}[2]{{%
\expandafter\ifx\csname l@#1\endcsname\relax
\typeout{** WARNING: IEEEtran.bst: No hyphenation pattern has been}%
\typeout{** loaded for the language `#1'. Using the pattern for}%
\typeout{** the default language instead.}%
\else
\language=\csname l@#1\endcsname
\fi
#2}}
\providecommand{\BIBdecl}{\relax}
\BIBdecl

\bibitem{Perona91}
P.~{Perona}, ``Deformable kernels for early vision,'' in \emph{1991 IEEE
  Computer Society Conference on Computer Vision and Pattern Recognition
  (CVPR)}, 1991, pp. 222--227.

\bibitem{GM95}
G.~M. {Haley} and B.~S. {Manjunath}, ``Rotation-invariant texture
  classification using modified gabor filters,'' in \emph{1995 International
  Conference on Image Processing (ICIP)}, vol.~1, 1995, pp. 262--265 vol.1.

\bibitem{T20}
T.~N. {Tan}, ``Rotation invariant texture features and their use in automatic
  script identification,'' \emph{IEEE Transactions on Pattern Analysis and
  Machine Intelligence}, vol.~20, no.~7, pp. 751--756, 1998.

\bibitem{YTang19}
Y.~{Tang} and X.~{Wu}, ``Salient object detection using cascaded convolutional
  neural networks and adversarial learning,'' \emph{IEEE Transactions on
  Multimedia}, vol.~21, no.~9, pp. 2237--2247, 2019.

\bibitem{HXiao12}
H.~{Xiao}, J.~{Feng}, Y.~{Wei}, M.~{Zhang}, and S.~{Yan}, ``Deep salient object
  detection with dense connections and distraction diagnosis,'' \emph{IEEE
  Transactions on Multimedia}, vol.~20, no.~12, pp. 3239--3251, 2018.

\bibitem{Freeman91}
W.~Freeman and E.~Adelson, ``The design and use of steerable filters,''
  \emph{IEEE Transactions on Pattern Analysis and Machine Intelligence,},
  vol.~13, pp. 891--906, 10 1991.

\bibitem{Simoncelli96}
E.~Simoncelli and H.~Farid, ``Steerable wedge filters for local orientation
  analysis,'' \emph{IEEE transactions on image processing}, vol.~5, pp.
  1377--82, 02 1996.

\bibitem{Taco16}
S.~C. Taco and W.~Max, ``Steerable cnns,'' in \emph{2016 International
  Conference on Learning Representations (ICLR)}, 2016.

\bibitem{Zhou17}
Y.~Zhou, Q.~Ye, Q.~Qiu, and J.~Jiao, ``Oriented response networks,'' in
  \emph{2017 IEEE Conference on Computer Vision and Pattern Recognition
  (CVPR)}, 2017, pp. 4961--4970.

\bibitem{S18}
S.~{Luan}, C.~{Chen}, B.~{Zhang}, J.~{Han}, and J.~{Liu}, ``Gabor convolutional
  networks,'' \emph{IEEE Transactions on Image Processing}, vol.~27, no.~9, pp.
  4357--4366, 2018.

\bibitem{Lee99}
D.~D. Lee and H.~S. Seung, ``Learning the parts of objects by nonnegative
  matrix factorization,'' \emph{Nature}, vol. 401, no. 6755, pp. 788--791,
  1999.

\bibitem{2003Optimally}
D.~L. Donoho and M.~Elad, ``Optimally sparse representation in general
  (nonorthogonal) dictionaries via l~1 minimization,'' \emph{Proceedings of the
  National Academy of Sciences of the United States of America}, vol. 100,
  no.~5, pp. 2197--2202, 2003.

\bibitem{96Ol}
B.~Olshausen and D.~Field, ``Emergence of simple-cell receptive field
  properties by learning a sparse code for natural images,'' \emph{Nature},
  vol. 381, pp. 607--609, 07 1996.

\bibitem{06Elad}
M.~{Elad} and M.~{Aharon}, ``Image denoising via learned dictionaries and
  sparse representation,'' in \emph{2006 IEEE Computer Society Conference on
  Computer Vision and Pattern Recognition (CVPR)}, vol.~1, 2006, pp. 895--900.

\bibitem{08JMairal}
J.~{Mairal}, M.~{Elad}, and G.~{Sapiro}, ``Sparse representation for color
  image restoration,'' \emph{IEEE Transactions on Image Processing}, vol.~17,
  no.~1, pp. 53--69, 2008.

\bibitem{YZ19}
Y.~{Zhou}, A.~{Mao}, S.~{Huo}, J.~{Lei}, and S.~{Kung}, ``Salient object
  detection via fuzzy theory and object-level enhancement,'' \emph{IEEE
  Transactions on Multimedia}, vol.~21, no.~1, pp. 74--85, 2019.

\bibitem{GM20}
G.~{Ma}, C.~{Chen}, S.~{Li}, C.~{Peng}, A.~{Hao}, and H.~{Qin}, ``Salient
  object detection via multiple instance joint re-learning,'' \emph{IEEE
  Transactions on Multimedia}, vol.~22, no.~2, pp. 324--336, 2020.

\bibitem{08Lee}
H.~Lee, C.~Ekanadham, and A.~Ng, ``Sparse deep belief net model for visual area
  v2,'' in \emph{Advances in Neural Information Processing Systems (NIPS)},
  vol.~20, 2008, pp. 873--880.

\bibitem{Hoyer04}
P.~O. Hoyer, ``Non-negative matrix factorization with sparseness constraints,''
  \emph{Journal of Machine Learning Research}, vol.~5, no.~3, 2004.

\bibitem{2019Sparse_Prin}
``Sparse principal component analysis,'' \emph{Journal of Computational and
  Graphical Statistics}.

\bibitem{10Gregor}
K.~Gregor and Y.~Lecun, ``Learning fast approximations of sparse coding,'' in
  \emph{2010 International Conference on International Conference on Machine
  Learning (ICML)}, 08 2010, pp. 399--406.

\bibitem{07BSch}
B.~{Scholkopf}, J.~{Platt}, and T.~{Hofmann}, ``Efficient sparse coding
  algorithms,'' in \emph{Advances in Neural Information Processing Systems
  (NIPS)}, 2007, pp. 801--808.

\bibitem{Y.X13}
Y.~{Wang}, H.~{Xu}, and C.~{Leng}, ``Provable subspace clustering: When lrr
  meets ssc,'' \emph{IEEE Transactions on Information Theory}, vol.~65, no.~9,
  pp. 5406--5432, 2019.

\bibitem{V.M15}
V.~M. {Patel}, H.~{Van Nguyen}, and R.~{Vidal}, ``Latent space sparse and
  low-rank subspace clustering,'' \emph{IEEE Journal of Selected Topics in
  Signal Processing}, vol.~9, no.~4, pp. 691--701, 2015.

\bibitem{46Gabor}
D.~Gabor, ``Theory of communication,'' \emph{Electrical Engineers Part I
  General Journal of the Institution}, vol.~93, pp. 429--459, 01 1946.

\bibitem{10Baochang}
{Baochang Zhang}, {Yongsheng Gao}, {Sanqiang Zhao}, and {Jianzhuang Liu},
  ``Local derivative pattern versus local binary pattern: Face recognition with
  high-order local pattern descriptor,'' \emph{IEEE Transactions on Image
  Processing}, vol.~19, no.~2, pp. 533--544, 2010.

\bibitem{HL14}
H.~{Liu}, G.~{Yang}, Z.~{Wu}, and D.~{Cai}, ``Constrained concept factorization
  for image representation,'' \emph{IEEE Transactions on Cybernetics}, vol.~44,
  no.~7, pp. 1214--1224, 2014.

\bibitem{J.F14}
J.~{Feng}, Z.~{Lin}, H.~{Xu}, and S.~{Yan}, ``Robust subspace segmentation with
  block-diagonal prior,'' in \emph{2014 IEEE Conference on Computer Vision and
  Pattern Recognition (CVPR)}, 2014, pp. 3818--3825.

\bibitem{DA01}
D.~A.~V. Dyk and X.-L. Meng, ``The art of data augmentation,'' \emph{Journal of
  Computational and Graphical Statistics}, vol.~10, no.~1, pp. 1--50, 2001.

\bibitem{Laptev16}
D.~Laptev, N.~Savinov, J.~M. Buhmann, and M.~Pollefeys, ``Tipooling:
  transformation-invariant pooling for feature learning in convolutional neural
  networks,'' in \emph{2016 IEEE Conference on Computer Vision and Pattern
  Recognition (CVPR)}, 2016.

\bibitem{Wei17}
Y.~Wei, ``Deformable convolutional networks,'' in \emph{2017 European
  Conference on Computer Vision (ECCV)}, 2017.

\bibitem{Jaderberg15}
M.~Jaderberg, K.~Simonyan, A.~Zisserman, and K.~Kavukcuoglu, ``Spatial
  transformer networks,'' in \emph{Advances in Neural Information Processing
  Systems (NIPS)}, 2015, pp. 2017--2025.

\bibitem{Zagoruyko16}
S.~Zagoruyko and N.~Komodakis, ``Wide residual networks,'' \emph{arXiv}, 2016.

\bibitem{He06}
K.~He, X.~Zhang, S.~Ren, and J.~Sun, ``Deep residual learning for image
  recognition,'' in \emph{2016 IEEE Conference on Computer Vision and Pattern
  Recognition (CVPR)}, 06 2016, pp. 770--778.

\bibitem{CL18}
C.~{Lu}, J.~{Feng}, S.~{Yan}, and Z.~{Lin}, ``A unified alternating direction
  method of multipliers by majorization minimization,'' \emph{IEEE Transactions
  on Pattern Analysis and Machine Intelligence}, vol.~40, no.~3, pp. 527--541,
  2018.

\bibitem{A.B09}
A.~Beck and M.~Teboulle, ``A fast iterative shrinkage-thresholding algorithm
  for linear inverse problems,'' \emph{SIAM Journal on Imaging Sciences},
  vol.~2, no.~1, pp. 183--202, 2009.

\bibitem{J.C10}
J.~Cai, E.~Candes, and Z.Shen, ``A singular value thresholding algorithm for
  matrix completion,'' \emph{SIAM journal on control and optimization},
  vol.~20, no.~4, pp. 1956--1982, 2010.

\bibitem{ZL18}
Z.~{Li}, J.~{Tang}, and X.~{He}, ``Robust structured nonnegative matrix
  factorization for image representation,'' \emph{IEEE Transactions on Neural
  Networks and Learning Systems}, vol.~29, no.~5, pp. 1947--1960, 2018.

\bibitem{2013Robust1}
G.~Liu, Z.~Lin, S.~Yan, J.~Sun, Y.~Yu, and Y.~Ma, ``Robust recovery of subspace
  structures by low-rank representation,'' \emph{IEEE Transactions on Pattern
  Analysis and Machine Intelligence}, vol.~35, no.~1, pp. 171--184, 2013.

\bibitem{Lin14}
M.~Lin, Q.~Chen, and S.~Yan, ``Network in network,'' in \emph{2014
  International Conference on Learning Representations (ICLR)}, 2014.

\bibitem{Netzer11}
A.~C. A. B. B.~W. Y.~Netzer, T.~Wang and A.~Y. Ng, ``Reading digits in natural
  images with unsupervised feature learning,'' in \emph{Advances in Neural
  Information Processing Systems (NIPS)}, 2011.

\end{thebibliography}

\end{document}